# Preliminaries to artificial consciousness: a multidimensional heuristic approach.

**K. Evers[1, *], M. Farisco[1,2,*], R. Chatila,[3] B. D. Earp,[4,5] I. T. Freire[3], F. Hamker[6], E. Nemeth[3], P. F. M. J. Verschure[7], M. Khamassi[3]**

[1] Centre for Research Ethics and Bioethics, Uppsala University, Uppsala, Sweden
[2] Biogem Molecular Biology and Genetics Research Institute, Ariano Irpino (AV), Italy
[3] Institute of Intelligent Systems and Robotics, Sorbonne University / CNRS, Paris, France
[4] Uehiro Centre for Practical Ethics, University of Oxford, Oxford, England
[5] Centre for Biomedical Ethics, National University of Singapore, Singapore
[6] Artificial Intelligence, Computer Science, Chemnitz University of Technology, Germany
[7] Alicante Institute of Neuroscience & Department of Health Psychology, Universidad Miguel Hernandez, Spain

* Shared first authorship

## Abstract

The pursuit of artificial consciousness requires conceptual clarity to navigate its theoretical and empirical challenges. This paper introduces a composite, multilevel, and multidimensional model of consciousness as a heuristic framework to guide research in this field. Consciousness is treated as a complex phenomenon, with distinct constituents and dimensions that can be operationalized for study and for evaluating their replication. We argue that this model provides a balanced approach to artificial consciousness research by avoiding binary thinking (e.g., conscious vs. non-conscious) and offering a structured basis for testable hypotheses. To illustrate its utility, we focus on "awareness" as a case study, demonstrating how specific dimensions of consciousness can be pragmatically analyzed and targeted for potential artificial instantiation. By breaking down the conceptual intricacies of consciousness and aligning them with practical research goals, this paper lays the groundwork for a robust strategy to advance the scientific and technical understanding of artificial consciousness.

## 1. Introduction

The possibility of artificial consciousness (roughly, subjective awareness in a human-designed artificial system) has been assumed within certain theoretical frameworks [1]. However, whether artificial consciousness is indeed theoretically possible, much less empirically feasible, is not self-evident, and neither proposition should be taken for granted. Nevertheless, with the rapid progression of relevant technologies, the prospect of producing artificial forms of consciousness is gaining traction in both scientific and public debates, eliciting different and sometimes opposing reactions [2]. The two extremes range from an optimistic enthusiasm emphasising the unavoidable emergence of artificial consciousness on the one hand [3], to a pressing call for caution, on the other hand (e.g., [4]), occasionally mixed with scepticism about the feasibility of any attempt to artificially recreate consciousness (see, e.g., [5]). A number of alternative views lay in between, each leaning *pro* or *contra* the conceivability, plausibility, feasibility, and (not least) desirability of artificial consciousness on the basis of various theoretical, scientific and socio-ethical arguments [6-15].

To reflect on these issues, researchers from different fields have used a number of distinct approaches. These have included: starting from leading scientific theories of consciousness in order to infer relevant indicators of consciousness and eventually check their applicability to current artificially intelligent (AI) systems [6]; theoretically reflecting on the necessary and sufficient conditions for consciousness and their possible instantiation in such systems [7]; philosophically and critically analysing the applicability of notions like intelligence and consciousness to technological artefacts [8, 16]; performing ethical analysis of what the prospect of artificial consciousness, including synthetic phenomenology, would imply for either human subjects or AI systems themselves[9]; reflecting on what conscious machines may entail for society on a descriptive level [10]; identifying reliable indicators for artificial consciousness [17] and relevant tests [14, 18], including a relevant ethical analysis [15]; reflecting on the risks related to the possible confusion about the sentience of AI systems [19]; and taking biological consciousness and its relation with the brain as a reference to critically evaluate the feasibility of artificial consciousness [13, 20, 21]. Therefore, the discussion on artificial consciousness is quite multifaceted and includes different complementary and partly overlapping aspects that are not easy to summarise within a unitary perspective.

The topic of creating artificial consciousness is controversial in part because it deals with highly sensitive issues and much is at stake: consciousness is a notion extremely prone to anthropocentric and anthropomorphic interpretations, and attributing it to other systems (whether biological or artificial) may raise different reactions, either defensive (e.g., arguing consciousness is human-specific, or only shared with other animals) or readily embracing the idea that artificial systems could be conscious [22]. These reactions are sometimes triggered by lack of clarity; notably, both disproportionate optimism or/and disproportionate scepticism about artificial consciousness can be due to a lack of clarity about what is actually at stake [23]. Accordingly, a more precise and fine-grained understanding of consciousness, including a more analytical identification of its specific components and dimensions is useful if not necessary for pursuing a balanced and realistic discussion of artificial consciousness, whether through the delineation of a route towards its realisation, or through the identification of obstacles that may either be temporary or fundamentally insurmountable. The need for such a conceptual elaboration also illustrates that the study of consciousness is still in a pre-scientific phase (in the Kuhnian sense of the coexistence of different theories each claiming its own scientific statute)[24], which asks for further modesty in our approach.

To advance in this debate, we consider it crucial to proceed on the basis of a careful and balanced theoretical reflection informed by empirical data. Such a theoretical analysis should initially be as conceptually unbiased (e.g., ideologically, politically, scientifically, and philosophically) and neutral as possible regarding the core questions of the conceivability, plausibility, feasibility, and desirability of artificial consciousness.

The primary goal of this paper is to propose a **composite, multilevel, and multidimensional model of consciousness** to clarify conceptual issues surrounding artificial consciousness. In fact, consciousness, with its multifaceted nature, presents unique difficulties for operationalization and empirical study. This paper aims to provide a structured approach to these challenges by introducing a heuristic framework that can guide research in this area. The proposed model treats consciousness as a composite phenomenon, characterized by different constituents, dimensions, and levels. This multidimensional approach avoids oversimplified binary categorizations (e.g., conscious vs. non-conscious) and instead provides a nuanced perspective that captures the spectrum of conscious states. By doing so, the model serves as both a theoretical tool for clarifying key concepts and an empirical guide for developing testable hypotheses.
A central aspect of this paper is the application of the proposed model to specific research questions. We focus on "awareness" as an illustrative case study, demonstrating how the model can inform the analysis and potential artificial realization of this constituent of consciousness. Awareness was selected for its clinical relevance, conceptual accessibility, and empirical tractability, making it an ideal example to showcase the model's utility.
This paper is structured as follows: first, we outline some logical and conceptual conditions necessary for studying artificial consciousness (section 2). We then introduce key terminological distinctions to establish a clear conceptual framework (section 3). Next, we detail the composite, multidimensional, and multilevel model, followed by a focused examination of awareness as a case study (section 4). Finally, we discuss the implications of this approach for advancing artificial consciousness research and propose directions for future inquiry (section 5). By integrating conceptual clarity with practical applicability, this paper aims to lay the groundwork for a balanced and effective research strategy in artificial consciousness.

## 2. Logical conditions for the theoretical analysis of artificial consciousness

For theoretical reflection on artificial consciousness to be effective, it must be characterized by *analytical clarity* and *logical coherence*. Analytical clarity refers to the needed unambiguous explanation of the terms invoked and their reciprocal connections and requires consistency in the use of terminology. Importantly, the different meanings of the same terms in different contexts (e.g., scientific vs. public debates) should be acknowledged and carefully accounted for in the communication of scientific and technological achievements concerning artificial consciousness, both in general and in any of its specific forms or components (e.g., awareness) in particular. This is especially true for consciousness, which is a highly sensitive issue: as the long discussion about animal consciousness illustrates [25-27], misunderstandings can cause disproportionate reactions, which may arise from passionate and ideologically driven positions rather than from empirically informed, rational reflection [22]. Also, finding a shared, overarching definition of consciousness is a very challenging and still open task, and for our purposes here it may be sufficient to agree on a working or stipulative definition for providing more clarity and consistency to the discussion about its artificial development.

Ensuring logical coherence is paramount in scholarly discourse, particularly to avoid common logical traps and fallacies. One such fallacy we identify in the field of artificial consciousness is the **analytical fallacy**. This occurs when one attempts to derive new empirical findings directly from a presupposed theory that either does not meet the falsifiability criterion (even in its moderate or pragmatic forms, acknowledging that strict falsifiability may be challenging) [28] or lacks sufficient empirical validation. The analytical fallacy specifically refers to the inappropriate conflation of analytical (linguistic or conceptual) statements with **synthetic** (empirical) statements. Analytical statements are true by virtue of their meanings and logical form, whereas synthetic statements are contingent on empirical evidence.
For instance, consider the Integrated Information Theory (IIT) [29], which posits that the degree of consciousness corresponds to the level of integrated information within a system. If one empirically measures aspects such as information integration and differentiation in a system and concludes that these measures indicate the presence of consciousness above a certain threshold, this reasoning may commit the analytical fallacy. The deduction of empirical evidence (consciousness) directly from theoretical premises (integrated information) without independent empirical validation can lead to logical inconsistencies.

Drawing conclusions from a theory that lacks independent empirical support is not inherently fallacious. However, when a theory claims to offer empirical discoveries and these discoveries are deduced solely from the theory's premises, it risks falling into circular reasoning. This circularity presupposes the very existence of what it aims to prove—in this case, artificial consciousness—**without clearly outlining the specific theoretical frameworks and experimental contexts necessary for empirical validation**. Without such specifications, the claims remain abstract and untestable, undermining the robustness of the argument.

The analytical fallacy eventually results in a circular thinking, surreptitiously presuming what should be proven (i.e., that artificial consciousness is actually possible or even real) rather than specifying in which framework and context (if any) artificial consciousness may be empirically possible or real. This leads to "ironic science" (a concept originally coined by Horgan, [30]) which transcends falsification [31], both in principle and *de facto*.

It is important to note that empirical plausibility (let alone actuality) cannot be inferred from theoretical possibility and logical conceivability: empirical considerations must be added to justify any such inference. The fact that something can be logically conceived (an extremely large set of possibilities, we may note) is not a sufficient condition for its empirical possibility, plausibility and actuality: additional factors must be taken into account. In the case of artificial consciousness, these would include, among other things, the availability of necessary technology, a sufficient understanding of the only knows physical instantiation (e.g., relevant biological processes) of consciousness to possibly emulate, the capacity to translate the principles underlying those processes into technological systems, and empirical indicators supporting the presence of consciousness in an artificial entity.

## 3. Relevant terminological distinctions in consciousness studies

In addition to the aforementioned logical conditions, a preliminary terminological clarification is necessary for an effective reflection about artificial consciousness. In particular, the notion of consciousness needs clarification, since it is open to different and sometimes not fully compatible or even incompatible understandings.

In the following we scrutinize some concepts that have been elaborated within the literature on consciousness and that are instrumental to introduce our proposal for advancing in the debate about artificial consciousness in particular. Therefore, we focus only on selected concepts, without any presumption of covering all the conceptual complexity of consciousness.

### 3.1 Access vs. phenomenal consciousness

A classical distinction within the philosophy of consciousness is that between access and phenomenal consciousness. These are usually presented as two different concepts and are theorised to correspond to two different forms of consciousness (See [32, 33] for a critical view of the actual distinction between access and phenomenal consciousness, and [34] for a supportive view about this distinction).

As defined by Ned Block, access consciousness refers to the interaction between different mental states, particularly the availability of one state's content for use in another, for example, for purposes of reasoning or rationally guiding capabilities like speech and action (i.e., the cognitive availability of information); whereas phenomenal consciousness is the subjective feeling of a particular experience, "what it is like to be" in a particular state (i.e., including experiences of perceptions which are not cognitively accessed) [35, 34]. More specifically, access consciousness relies on information provided by different cognitive processes mediating functions like working memory, verbal report and motor behaviour [36], while phenomenal consciousness refers to the subjective experience of the conscious subject characterised by a specific point of view.

Interestingly, this distinction is not universally accepted [37-39, 33, 40]. Some researchers deny the existence of phenomenal consciousness as a specific form separated from access consciousness and propose to replace phenomenal consciousness with the differentiation of **levels of conscious access**. On this account, the subject would be able to access the phenomenal contents, but not always to verbally report them [41]. Therefore, contrary to what some philosophers, including Block, have argued [42-44], these researchers do not think that phenomenal consciousness can overflow (i.e., contain more information than) access consciousness[45], but rather distinguish between access consciousness and reportability, and eventually reduce phenomenal consciousness to access consciousness [32].

The discussion is still open about this proposed rejection of the dichotomy between access and phenomenal consciousness, including the fact that, depending on some of the more specific interpretations, this rejection may greatly diminish the number of animal species that can be considered conscious. For one thing, if phenomenal consciousness is reduced to access consciousness, and this is limited to higher cognitive functions, less cognitively complex animal species may be excluded a priori from the realm of consciousness.

Moreover, requiring reportability (to humans) is arguably anthropocentric and speciesist: it presumes (the necessity of) human language capacities and/or human conceptualisations, or understanding of actions. Other species may have highly developed **forms of consciousness** that we have (or maybe lack), that can be reportable in a sense to their kin but not to us (we have for example an extremely poor level and range of perception, so most sensory expressions of other animals are as hidden from us as colours are to the blind and music to the deaf). At the other extreme, conflating phenomenal and access consciousness and requiring reportability may lead to both false positives and false negatives in the case of artificial systems, which are increasingly able to imitate highly evolved cognitive and communication abilities.

Interestingly, this reduction of phenomenal consciousness to access consciousness is counterbalanced by a wide tendency to identify consciousness *tout court* with phenomenal consciousness, at least in a morally relevant sense [46], or to affirm the necessity of phenomenal consciousness for access consciousness to be possible. According to these perspectives, the real question about a technically highly advanced artificial entity, e.g., AI is whether it is capable of developing an artificial form of subjective experience. This is by far the most interesting version of the question whether an artificial system could become conscious, but also, as we will detail below, the most challenging.

How the relationship between access and phenomenal consciousness is conceived (e.g., as an epistemic strategy or a distinction between concepts rather than a functional description or a distinction between two actual forms of consciousness) also has implications for the issue of artificial consciousness. In fact, if they are two different forms of consciousness, it is theoretically possible (even if it remains an open question empirically) that access and phenomenal consciousnesses are dissociable, and that an AI system may possess only one of the two forms. Even if in principle this would be a consistent conclusion, given the above mentioned tendency to identify consciousness *tout court* with phenomenal consciousness, it would definitely not be sufficient to deal with the elephant in the room (i.e., could an AI system subjectively experience anything?).

### 3.2 Level vs. content of consciousness

For clinical purposes, an operational distinction between two components of consciousness has gained traction in recent years: *level* and *content* of consciousness are identified as the two axes along which it is possible to assess consciousness, more specifically to both quantify it and rate the capacity of the subject to consciously perceive particular objects. Accordingly, level and content of consciousness are identified with wakefulness and awareness respectively, and consciousness is graded within a two-dimensional framework going from coma to sleep to conscious wakefulness [47]. Also, this two-dimensional understanding of consciousness has been used to characterize ictal alterations, (i.e., changes or symptoms that occur during a seizure -ictal phase- in epilepsy)[48].

Recently, some have proposed an extended axis including, e.g., psychedelic experiences as being even 'higher' levels of consciousness than 'mere' conscious wakefulness, but this view is controversial [49].

In the clinical context, awareness is defined as the content-related component of conscious experience, in addition to wakefulness (or the level of vigilance). In its minimal clinical definition, awareness is the capacity of the subject to process information, store it in short-term memory, and possibly intentionally retrieve it from long-term memory if needed. In fact, this minimal clinical characterization of awareness does not clearly identify minimal conditions for distinguishing it from non-conscious operations [50-52]. We propose that the intentional use of information for achieving goals stands as a minimal necessary condition[1] for aware processing. Importantly, this intentionality may be taken to suggest also non-zero subjective experience.

We may say that as a content representation (i.e., awareness), consciousness is intentional (i.e., directed to something), while level and state (i.e., wakefulness) refer to the preliminary capacity for this representational process, and they may or may not eventually correlate with aware consciousness. The clinical case of vegetative state/unresponsive wakefulness syndrome (VS/UWS), defined as wakefulness without awareness [52, 47], as well as dream states, where the subject is aware but unawake [53], are illustrative of the potential dissociation between these two components of consciousness (i.e., awareness and wakefulness).

Thus, there are two conditions for a system to actually possess aware consciousness: the capacity for processing information combined with the capacity for intentionally using it (i.e., identifying and making use of affordances in the surrounding environment). If in addition to intentional action as defined above a system is also able to attach a value to the processed information, then it is capable of subjective experience; it is like something to be that system.

In conclusion, in the traditional clinical understanding of consciousness, this results from the combination of awareness with wakefulness.

The clinical two-tier view of consciousness has been criticized, and a multidimensional model has been proposed as an alternative to it [54]. As we will explain with more details below, we agree that a multidimensional model of consciousness is a promising approach to explore the possibility of artificial consciousness, even if we think that the concept of level may be applied also to each dimension of consciousness, going beyond the level/content dichotomy, as we will discuss later.

### 3.3 The conceptual prism of consciousness

The selected notions and approaches presented above are not comprehensive but sufficient to introduce the high level of controversy surrounding consciousness, including its scientific understanding and its possible replication in AI systems.

To summarise (see Fig. 1), the term consciousness may refer to cognitive information processing or to subjective experience (even if it is not unanimously accepted that access and phenomenal consciousnesses actually are two distinct forms rather than two different concepts of consciousness), to the level of the subject's consciousness (e.g., awake vs. unawake) or to the content of the conscious experience (i.e., awareness). Both the state/level (wakefulness) and the content (awareness) are considered as two fundamental components of consciousness, at least in the clinical context.

---

[1] We qualify intentionality (i.e., a goal-oriented action) as the minimal necessary condition for awareness because we think that the intentional use of information for achieving specific goals is crucial for distinguishing between aware and unaware cognition. At the same time, we do not exclude that other dimensions of awareness as listed hereafter are also present, even if they are not minimally necessary.

Importantly, the conscious cognitive content resulting from the capacity to process information (i.e., to be aware, either in awake or non awake state) is not the same as the conscious subjective experience, including subjective perceptual experience (i.e., sensation [55]), which connects to the phenomenal, subjective form of consciousness (i.e., what it is like to be in a specific state, [35, 56, 34]).

Also, the relation between consciousness and the self is important to clarify. Self-consciousness is one possible component of conscious experience, but a robust and reflective self-perception is not necessary for conscious experience in general. In fact, self-consciousness is multilevel [57]. It is possible for a subject to be aware and also to have a minimal phenomenal experience [58] even if lacking a strong reflective sense of self, that is a self-oriented meta-cognitive capacity, both among non-human animals (e.g., dolphins, octopuses, crows, or bonobos) and humans (e.g., infants, and adults in psychedelic experiences) [17, 27, 59-67].

Thus, the definition of consciousness is a most challenging task. A universal agreement about it is hardly achievable, mainly due to the myriad different theoretical models and related definitions [68]. These distinct theories are not necessarily commensurable, or they can be so in different ways, and the usefulness of common denominators in differentiating, integrating and testing hypotheses has recently been analysed [20]. There are also attempts to elaborate a unifying model of consciousness [69, 70, 21], to identify a reliable measure for an empirical comparison among different theories [71], or even to implement an adversarial collaboration among different theories [72], yet the question is open about how to advance towards a more mature science of consciousness, including more robust agreement about its definition [73]. A possible strategy in this direction is considering the different theories as complementary rather than adversarial or alternative to each other [74].

We suggest that it is not necessary to agree about an overarching, general definition of consciousness for reflecting about the possibility and plausibility of artificial consciousness arising. In fact, there are two alternatives: to agree on a working or a stipulative definition in order to advance towards a sufficient level of agreement and mutual understanding, especially between scientific researchers and people from other disciplines (e.g., social and political science, ethics, philosophy), and also from the general public; or to agree on what are at least some of the general features that characterise consciousness, including specific abilities that are facilitated by consciousness, beyond the specific theoretical stance one endorses. The latter approach seems especially useful for identifying criteria or indicators that may facilitate the attribution of conscious capacities to other human and nonhuman agents, including artificial systems.

This approach is consistent with the "theory-light" approach as recently described by Birch as a methodology that relies on a minimal commitment about the relation between (phenomenal) consciousness and cognition, so that it does not subscribe to any specific theory of consciousness [75]. The core hypothesis is that conscious perception of a stimulus facilitates a cluster of cognitive abilities in relation to that stimulus so that their presence indicates the occurrence of conscious perception. In the following, we follow this direction, introducing some components and dimensions of consciousness that can arguably be considered characteristic to it. Then among the different concepts reviewed above we take awareness and the question about its artificial replication as a case-study to illustrate how a composite, multidimensional, and multilevel model of consciousness may inspire an effective and balanced research strategy on artificial consciousness, which may result in testable hypotheses.

Building on these clarified terms, we introduce a composite model to address the challenges posed by the multifaceted nature of consciousness.

## 4. A composite and multidimensional heuristic model of consciousness

In the light of the different meanings attributed to consciousness as introduced above, it is reasonable to infer that consciousness presents different **constituents** (i.e., states, forms, components, and dimensions), as reflected in the different senses of the term (Fig. 1). For instance, wakefulness and awareness are considered as two fundamental **components** (i.e., building blocks) of consciousness, at least in the clinical context. The same with two fundamental forms (or kinds) of consciousness like access consciousness and phenomenal consciousness, provided that one agrees with considering them as two concepts corresponding to two distinct phenomena, and that one especially agrees with the actual existence of the latter. Moreover, each constituent of consciousness (both as a cognitive appraisal and as a phenomenological state) is arguably **multidimensional** [76-78]. In other words, we think that the description of **consciousness as a composite, multidimensional, and multilevel feature** may work as a synthetic notion that summarizes its prismatic nature. In addition to this conceptual utility, this kind of description has the advantage to make consciousness testable, because it identify specific objects and functions to be possibly tested in the laboratory.

More specifically, Bayne, Hohwy, and Owen [54] argue that global states of consciousness manifest themselves in multiple ways, and that the notion of levels should be replaced by that of dimensions of consciousness to properly describe it. The central thesis is that global states of consciousness are not gradable along one dimension, but rather distinguished along different dimensions. More specifically, they introduce two main families of **consciousness' dimensions**: content-related and functional.

The first family includes, for instance, gating of conscious content (e.g., low-level features vs high-level features of an object). The second family includes, for instance, cognitive and behavioural control (i.e., the availability of conscious contents for control of thought and action). Along this line of analysis, Walter has recently proposed the following content-related dimensions: sensory richness, high-order object representation, semantic comprehension; and the following functional dimensions: executive functioning, memory consolidation, intentional agency, reasoning, attention control, vigilance, meta-awareness [79].

Other relevant reflections about consciousness' dimensions come from Birch et al., who with reference to animal consciousness specifically introduce the following dimensions [62]:

- *Perceptual-Richness*: any measure is specific to a sense modality, so there is no overall level of perceptual richness. Also, within a particular sense modality, perceptual richness can be resolved into different components (e.g., bandwidth, acuity, and categorization power for vision);
- *Evaluative-Richness*: affectively-based positive or negative valence which grounds decision-making. Also evaluative richness can be resolved into different components;
- *Integration at a time (unity)*: conscious experience is (usually) highly unified;
- *Integration across time (temporality)*: conscious experience takes the form of a continuous stream;
- *Self-consciousness (Selfhood)*: awareness of oneself as distinct from the world outside.

Dung and Newen have introduced additional dimensions, again with explicit reference to animal consciousness, but potentially relevant for AI consciousness as well [63]. Within the category "external representation", where they include Perceptual-Richness and Evaluative-Richness as defined by Birch et al., they add Evaluative-Intensity, defined as how strongly a subject feels the positive or negative valence of an object/experience, and the external diachronic and synchronic unity.

Within the category "self-representation", they add self-referred diachronic and synchronic unity, experience of agency (i.e., the ability to experience actions as voluntarily initiated and controlled), and the experience of ownership (i.e., the ability to perceive body parts as something personal rather than objects of the external world). Within the category "cognitive processing strategies" they introduce three new dimensions: reasoning (e.g., complex trains of thought and ability to reason on multiple domains), learning (e.g., trace conditioning), and abstraction (i.e., the ability to form and use high-level abstract forms that categorise specific sensory stimuli).

Irwin has recently proposed another approach to animals' consciousness dimensions: on the basis of a behavioural study of twelve animal species, he identified three kinds of behaviour (volitional, interactive, and egocentric), quantified their frequency, variety, and dynamism, and eventually represented them in a matrix indicative of the **consciousness profile** of the animal in question[80].

All these attempts are illustrative of a highly lively debate that promises further advancement toward a more fine-grained and analytical reflection about consciousness and its constituents. We leave open the question whether the abovementioned dimensions are really dimensions of consciousness rather than cognition. Also, the dimensions listed above cover some aspects of the prism of consciousness while others remain less or not considered. For instance, another category of dimensions that appears not adequately addressed so far is social-relational functions or representations [81] and dyadic interactions, which include dimensions or capacities such as theory of mind (i.e., the ability to anticipate through a model-based virtualization the behaviour of others, particularly when instrumental to fulfilling personal goals), strategic collaboration (i.e., collaborating with others because it is instrumental to fulfil shared goals, even if particular benefits will be eventually reduced as a consequence of such collaboration, or the benefit is not immediate but postponed), and altruistic (or "communal") [82] orientations or behaviour (i.e., proneness to share resources if others are detected as in need, even if this sharing does not produce any personal benefit or raises the risk of reducing personal wellbeing).

While the justification of considering the abovementioned dimensions as dimensions of consciousness, the detailed identification of further dimensions of consciousness, or the identification of novel families and/or categories thereof, are still an open issue [79, 83], the concept of *consciousness profiles* emerges as the spaces of experience delimited by different specific dimensions within one or more constituents of consciousness.

Accordingly, we can differentiate consciousness profiles not in terms of their overall levels along one and the same dimension, but rather with reference to the combination of the different dimensions that characterise them (See Fig. 2 for a speculative illustration of the comparison between human and non-human consciousness profiles). For instance, it may well be the case that the consciousness profile of a human subject has some content-related and functional dimensions (e.g., semantic comprehension and meta-awareness, respectively) more advanced than a non-human entity, while other content-related and functional dimensions (e.g., sensory richness and vigilance, respectively) may be less advanced. Also, it is possible that a non-human entity (either biological or artificial) has a consciousness profile which includes some dimensions that humans lack (e.g., echolocation). This does not mean that one overall conscious state is higher or lower than the other, but rather that it is differently shaped.

Therefore, the comparison between human, other animals, and potential artificial consciousnesses should be framed in terms of resemblances and differences along specific constituents and related dimensions rather than in terms of higher or lower levels along only one, overall constituent and/or dimension. In short, consciousness is a multifaceted reality (i.e., a prism), irreducible to one level of description.

To summarize, in a pragmatic discussion about artificial consciousness it seems useful to consider consciousness as a complex feature defined by different constituents which have different dimensions. In addition, the notion of 'level' can be understood in a way that is compatible with this composite and multi-dimensional view of consciousness. In principle, a level of consciousness may indicate the grade of the global state consciousness (i.e., a rank along the same scale) or the specific form of consciousness that the subject is capable of (i.e., a differentiation among more or less sophisticated forms of consciousness, where the sophistication results from the richness of particular dimensions). This second meaning is compatible with the framework depicted above.

In conclusion, since consciousness is a composite and multilevel concept, it is necessary for any attempt to replicate it to specify which specific constituent (i.e., states, forms, components, and dimensions) is the target. We think that these clarifications and distinctions pave the way for a new research strategy to artificial consciousness which does not restrict itself to binary thinking (conscious versus non-conscious systems), nor to a single unidimensional level of consciousness (minimal versus high level of consciousness)[84], but that pragmatically target specific constituents of consciousness in the attempt to replicate them.

Even if in principle this kind of approach may take either cognitive or phenomenal constituents of consciousness as its object, we are aware that if consciousness is identified with phenomenal consciousness, and this is considered as the only ethically relevant form of consciousness, then the multidimensional approach depicted above may not be capable of providing a definitive answer to the fundamental question whether the artificial system is conscious or not. This risk may be counterbalanced by two considerations. First, we may identify a particular combination between the different dimensions that results in an overall phenomenal state if beyond a minimal threshold. This is an attractive possibility, still to be explored, and beyond the scope of the present paper. Second, we propose that this multidimensional approach is the best strategy to avoid both hype and misplaced worries about artificial consciousness.

### 4.1 How the Heuristic Model can guide research

The composite, multidimensional, and multilevel view of consciousness described above can serve as a heuristic framework, functioning as a guiding hypothesis in the classical sense. It is designed to inspire empirical research while addressing common conceptual challenges, such as avoiding binary categorizations or vague theoretical assumptions. By systematically breaking down consciousness into its constituents and dimensions, the model provides a structured and pragmatic approach for evaluating artificial consciousness.

As an heuristic tool, this model could play a pivotal role in advancing research on artificial consciousness. It can encourage the formulation of hypotheses that are both operationalizable and testable, thereby fostering a structured and methodical approach to exploring consciousness in artificial systems. By introducing measurable dimensions and components, it avoids conceptual pitfalls like overgeneralization or ambiguous definitions. Instead of relying on simplistic benchmarks such as passing a Turing Test, which are subject to the so-called "gaming problem" [85] the model supports nuanced evaluations that reflect the complexity of consciousness and guide the identification of specific research targets, such as awareness.

In practice, this model supports the definition of measurable dimensions of consciousness, such as sensory richness, intentional agency, or evaluative dimensions, allowing researchers to focus on distinct aspects that may be instantiated in artificial systems. It also informs the design of experiments aimed at probing these dimensions, offering a framework for validating theoretical assumptions through empirical evidence. For instance, awareness—a central focus of the next section—illustrates how a constituent of consciousness can be framed in terms of testable attributes, such as its role in goal-directed information processing.

Furthermore, the model provides a foundation for establishing benchmarks that capture the intricate nature of consciousness. Moving beyond binary assessments of conscious versus non-conscious systems, the heuristic model emphasizes the need for multidimensional evaluation, reflecting the true complexity of consciousness. By integrating these practical implications, the model ensures that research remains conceptually clear, empirically rigorous, and aligned with meaningful objectives. In order to demonstrate the utility of the multidimensional view of consciousness as an heuristic model, we now analyze the artificial realisation of awareness as an illustrative case-study, which provides a concrete example of how to select one of the multiple meanings attributed to consciousness, make it testable, and elaborate relevant research programs towards its replication.

**Key terms of consciousness**

**State**
A subjective overall conscious condition

**Constituent**
Element that makes up consciousness as a larger whole considered beyond its specific kind (i.e., states, forms, components, and dimensions)

**Component**
Element that makes up consciousness as a larger whole considered in relation to its specific kind (e.g., wakefulness and awareness as different states of consciousness; phenomenal and access consciousness as different forms of consciousness, etc.)

**Dimension**
Specific element of a component of consciousness. Dimensions are in principle measurable/quantifiable and may be of two main families: content-related and functional.

**Profile**
Specific characterization of a state of consciousness resulting from the combination of its different components and related dimensions

**Level**
Rank or degree of a specific dimension of consciousness or of the profile of consciousness resulting from the combination of its particular dimensions

**Form**
More or less sophisticated profile of consciousness resulting from the richness of its particular dimensions

## 5. Awareness as a case study for artificial consciousness research

As mentioned earlier, the present paper does not aim to propose an overarching definition of consciousness, but instead seeks to clarify relevant logical conditions and to provide foundational conceptual clarifications for advancing the discussion on the theoretical plausibility and the technical feasibility of artificial consciousness. Central to this endeavor is the heuristic framework proposed above, which offers a structured approach to dissecting and operationalizing the complex phenomenon of consciousness. This framework serves as a basis for establishing concrete benchmarks to assess specific dimensions, degrees and profiles of artificial consciousness. As part of this logical and conceptual reflection, the notions of consciousness' constituents, components, dimensions, and profiles have been introduced to guide this systematic inquiry.

Against this background, we now focus on awareness as a particular constituent of consciousness that contemporary AI may – or may not – succeed in instantiating. Awareness has been selected as the illustrative case study for this paper because of its clinical relevance and its accessibility for operationalization in both biological and artificial contexts. As a fundamental component of consciousness, awareness is extensively studied in clinical and cognitive sciences, particularly in conditions like disorders of consciousness and the assessment of minimal consciousness states. In the clinical context, awareness is considered a fundamental component of conscious experience, in addition to wakefulness (or the level of vigilance). In fact, there are several sets of empirical data about the neuronal mechanisms of this constituent of consciousness [47, 86]. The artificial realisation of the information-processing dimensions of awareness seems (at least intuitively) less controversial than the artificial realisation of the dimension of subjective experience (cognition and action control appear more prone to computational interpretation and replication, specially beyond academic circles).

This makes awareness both conceptually approachable and empirically tractable, providing a concrete basis for exploring how specific aspects of consciousness could be realized in artificial systems. By focusing on awareness, we aim to demonstrate how the composite, multidimensional, and multilevel model introduced in this paper can guide research on artificial consciousness in a structured and pragmatic manner.

This may seem insufficiently ambitious or too modest an approach, but we consider this a pragmatic and reasonable strategy to handle the complexity of consciousness as summarised in the above sections. We also think that this approach is promising for advancing the discussion in a balanced and realistic manner. In other words, awareness is a relevant example of how the composite, multidimensional, and multilevel model of consciousness introduced in this paper may be translated in a specific and possibly testable attempt to replicate consciousness in artificial systems.

According to the presented multidimensional framework, to be qualified as conscious, the capacity to process, store, and retrieve information that characterises awareness as defined in the clinical context should present different levels of both the content-related and functional dimensions that shape the consciousness profiles. As specified above, among those dimensions we here assume that two conditions are minimally necessary for aware processing: the capacity to select relevant information and the capacity to intentionally use it for achieving desired goals.

This minimal definition of awareness is open to different potential technical implementations. In fact, in order to be considered as one dimension of awareness, information processing should be more sophisticated than a model-free phenomenon (i.e., it should be more than simple input-output processing) and can go from basic levels when an agent has the capacity for modelling internal and external states with a prevailing monitoring goal and limited prediction capability, to higher levels when these models are combined with the capacity to virtualize the world and to predict more distant future states [21, 87-89]. To be markers of awareness, these capacities for modelling and virtualization should be combined with the capacity to intentionally exploit them as part of a goal-directed behaviour. These requirements, combined with the multi-dimensional definition we pursued here imply that for consciousness to be realised in artificial systems, it must be considered as a feature of a control architecture. Such an architecture must link primary and higher-order forms of local and global consciousness and show their integration with systems of perception, motivation, emotion, cognition and action. An example architecture that satisfies these requirements has been proposed in [21].

A neuro-inspired reinforcement learning (RL) architecture for robot online learning and decision-making has recently been developed in order to provide the system with the capacity to dynamically and autonomously adapt its behaviour to external circumstances in order to achieve its goals [90]. The architecture combines model-based (MB) and model-free (MF) RL, and it also includes a meta-controller for arbitrating between them in order to maximize efficiency and to minimize computational costs. In fact, the MB strategy builds a model of the long-term effects of actions and "decides" how to act on the basis of this model, thus enabling flexible adaptation but with high computational costs. On the other hand, the MF strategy is less flexible but far less computationally costly. Switching between these two strategies, the system is eventually able to assess the relevance of processed information for achieving its goals. Whereas these actions do not *ipso facto* involve intentionality, it may nevertheless be the case that, in a sense, this architecture implements some elements of awareness as defined above. It selects information relevant for achieving its goals, and it is able to proceed to enact the most effective strategy. To repeat, although this functioning is effective and minimally autonomous (this autonomy being limited by the fact that the reward function has been externally defined by the human designers of the model), it would not be legitimate to infer therefrom that it is intentional, for there are no indications that the system acts upon an internal drive (e.g., willingness). Notwithstanding, a fundamental capacity for selecting information instrumental to a goal-oriented strategy may be sufficient for a reasonable attribution of at least an important dimension of awareness to the system [87]. The same applies to other examples from recent AI research, like the Adaptive Agent developed by DeepMind [91]. Also this system has the capacity for on-the-fly hypothesis-driven exploration, efficient exploitation of acquired knowledge, and adaptation to open-ended novel circumstances, displaying the capacity for a goal-oriented action and use of information.

Of course, there are important aspects that this minimal definition of awareness leaves open, including the role of reward-based expectation for awareness (e.g., how does the anticipation of reward impact the selection of information the system is actually aware of [36, 92]) as well as the possible connection between feedforward and feedback dynamics in the system [21]. Furthermore, the connection between awareness and general intelligence, as well as between awareness and understanding remain open to different interpretations.

Another aspect of awareness that has been revealed by clinical research is the dissociation between internal or self-awareness (i.e., relative to the self) and external or sensory awareness [93]. Significantly, different networks for each of them have been identified (midline fronto-parietal and lateral fronto-parietal networks, respectively)[94, 95]. This confirms that consciousness does not require or imply self-consciousness, and that in principle it is possible to develop an aware artificial system devoid of self-awareness.

The insights derived from examining awareness as a case study extend beyond this specific constituent of consciousness. They provide a framework for analyzing other dimensions of consciousness, such as self-consciousness, evaluative capacities, or social and relational aspects like theory of mind. Each of these dimensions, like awareness, can be systematically operationalized and tested within the proposed composite heuristic model. By applying the multidimensional approach across various aspects of consciousness, researchers can explore the spectrum of conscious phenomena in both biological and artificial systems. This approach not only highlights the versatility of the heuristic framework but also underscores its potential to foster a deeper and more nuanced understanding of the challenges and possibilities in the study of artificial consciousness.

## 6. Discussion

As previously noted, we do not presume to provide a definitive answer to the questions whether artificial awareness could arise or how likely this development is. The answers to both questions depend on the background theoretical framework as well as on the technology actually available. For instance, the Distributed Adaptive Control Theory of consciousness assumes that artificial awareness is at least theoretically possible, setting the ground for the technological attempt to translate this possibility in reality [21]. In other words, the possibility of artificial awareness is assumed as a working hypothesis, which plays the role of a heuristic program inspiring empirical work towards its validation.

Being a component of consciousness, awareness does not exhaust its semantic and functional complexity. Even if limited, we propose to take awareness as a specific case-study of the attempt to produce conscious capacities in AI systems because awareness appears open to a more intuitive understanding and to a wider conceptual consensus than the general and sometimes opaque notion of consciousness, and the empirical investigation of the specific cerebral underpinnings of awareness is quite advanced. Even if the fundamental question about the theoretical plausibility of artificial awareness is still open, we propose that focusing on awareness may avoid the risk of a conceptual impasse and proceed more pragmatically towards the realisation of selected aspects of biological consciousness.

There are several issues that the approach we have introduced above still leaves open. While all merit further investigation, we here provide an illustrative list, which is by no means exhaustive, but only for the sake of introducing further important points of analysis.

A fundamental issue is why to pursue artificial awareness (and artificial consciousness more generally) in the first place: What would be the resulting benefits and advantages, for instance for science, or society at large? A possibility is that by building artificial awareness we will eventually better understand biological consciousness [21] . Another possibility is that building artificial awareness will be a game-changer in AI. In fact, the capacity to build world models is arguably an important factor for the further advancement of AI [96] especially in the interaction and collaboration with other agents, artificial and biological [21, 97]). For instance, artificial awareness would allow AI to intentionally use the world models it develops, resulting in both a significant improvement of AI technology and an important impact on society. Moreover, being aware of the consequences of its actions on the physical and social world could help AI better inform humans about potential negative impacts on society, and help avoid them while favouring positive impacts, which could contribute to a better alignment of AI systems with human values [98]). On the other hand, it is by no means obvious that an aware AI would be positively inclined towards or even at all interested in human welfare. Either way the question is open what impacts artificial awareness could or would have, and they require to be closely followed and regulated where necessary.

Another fundamental issue concerns the nature of a hypothetical artificial awareness: Is embodiment necessary for it? In other words, does awareness require an embodied subject, embedded in a particular environment (i.e., *Umwelt*), in relation to which it develops and makes intentional use of models for satisfying its needs? And what do we mean by "body" in this context: is being physical enough? Would aware AI residing in internet connections count as embodied? To address these questions about embodiment, one has to consider the origin of the information that must flow from the real world through a sensory system to be processed, and an interpretation of this information must be made to produce action. Any affordance would have to be built on the possibility of physical action. This argues in favour of embodiment, but the possible forms of embodiment need to be clarified.

Another challenging issue is the possibility that consciousness emerges in AI organically along its development. The hypothesis of an organic emergence of consciousness raises both theoretical and ethical issues: how to define consciousness? Which specific form of consciousness may emerge in AI? How to identify it?

Finally, the issue of values emerges as very challenging. For biological organisms, awareness is intrinsically related to the capacity to evaluate the world, discriminating between what is good and what is bad [99-101], including a capacity for a form of subjective experience. Is it the same for a possible artificial awareness? Or would the absence or the different nature of what makes values and evaluation necessary in biological awareness (e.g., emotions, reward-systems, preferences) eventually allow an artificial non-evaluative awareness? If so, would this be desirable? Values of different kinds (e.g., moral, political, religious) have inspired both positive and negative actions in human history, so the question is open about the moral implications of an aware agent devoid of any value.

Overall, the composite, multidimensional, and multilevel approach that we described in this paper aims to provide a heuristic model (i.e., a working hypothesis inspiring empirical work towards its validation) in order to advance the debate about the theoretical plausibility and the technical feasibility of artificial consciousness. Basically, this approach highlights that consciousness is like a spectrum which has many facets that may be instantiated at different degrees, in combination with or independently from each other. The question about the possibility of engineering phenomenal forms of consciousness, in particular, raises the most debated issues among both experts in the field and the general public. The approach described above does not provide a definitive answer to this question, while in principle it is not incompatible with a clear-cut binary answer about AI consciousness. In fact, in the framework proposed here, if the level of any dimension is higher than zero, then we may conclude that the AI system in question has a non-zero consciousness. To clarify in a more analytical way which kind of consciousness it has is a different task. Addressing it requires agreeing about specific benchmarks and thresholds, which is beyond the scope of the present analysis.

## Conclusions

In this paper, we addressed key theoretical issues—both logical and conceptual—that are crucial for advancing the study of artificial consciousness. Our primary contribution is the proposal of a composite, multilevel, and multidimensional model of consciousness, which serves as a heuristic framework to clarify conceptual ambiguities, inspire empirical research, and guide the systematic exploration of artificial consciousness. This model moves beyond simplistic, binary approaches by emphasizing the spectrum of conscious phenomena and providing a structured basis for operationalization and testing.

We illustrated the utility of this framework through the focused examination of "awareness" as a case study. Awareness was selected for its clinical relevance, conceptual accessibility, and empirical tractability. By analyzing this constituent of consciousness, we illustrated how the multidimensional model facilitates the identification of measurable dimensions and attributes, such as goal-directed information processing and intentionality. This case study highlights how the heuristic multidimensional model of consciousness can be applied to specific aspects of consciousness, offering a practical and balanced approach for research in artificial systems.

The broader implications of this approach extend beyond awareness, offering a pathway to explore other dimensions of consciousness, including self-consciousness, evaluative capacities, and social-relational aspects like theory of mind. By providing a coherent framework, the proposed model enables researchers to systematically investigate these dimensions, fostering a deeper understanding of the challenges and possibilities in replicating consciousness in artificial systems.

While the technical feasibility of replicating awareness or other forms of consciousness in AI systems remains an open empirical question, and the theoretical plausibility of replicating phenomenal consciousness continues to be debated, the proposed model lays the groundwork for addressing these issues. Achieving artificial consciousness will likely require a system-oriented architecture perspective, capable of integrating various dimensions and components into a coherent whole.

Despite these unresolved challenges, our framework provides a promising methodological foundation for advancing the discussion in a balanced, informed, and empirically grounded manner. By bridging theoretical clarity with practical applicability, it opens new avenues for research and supports the development of robust strategies for exploring the complex domain of artificial consciousness.

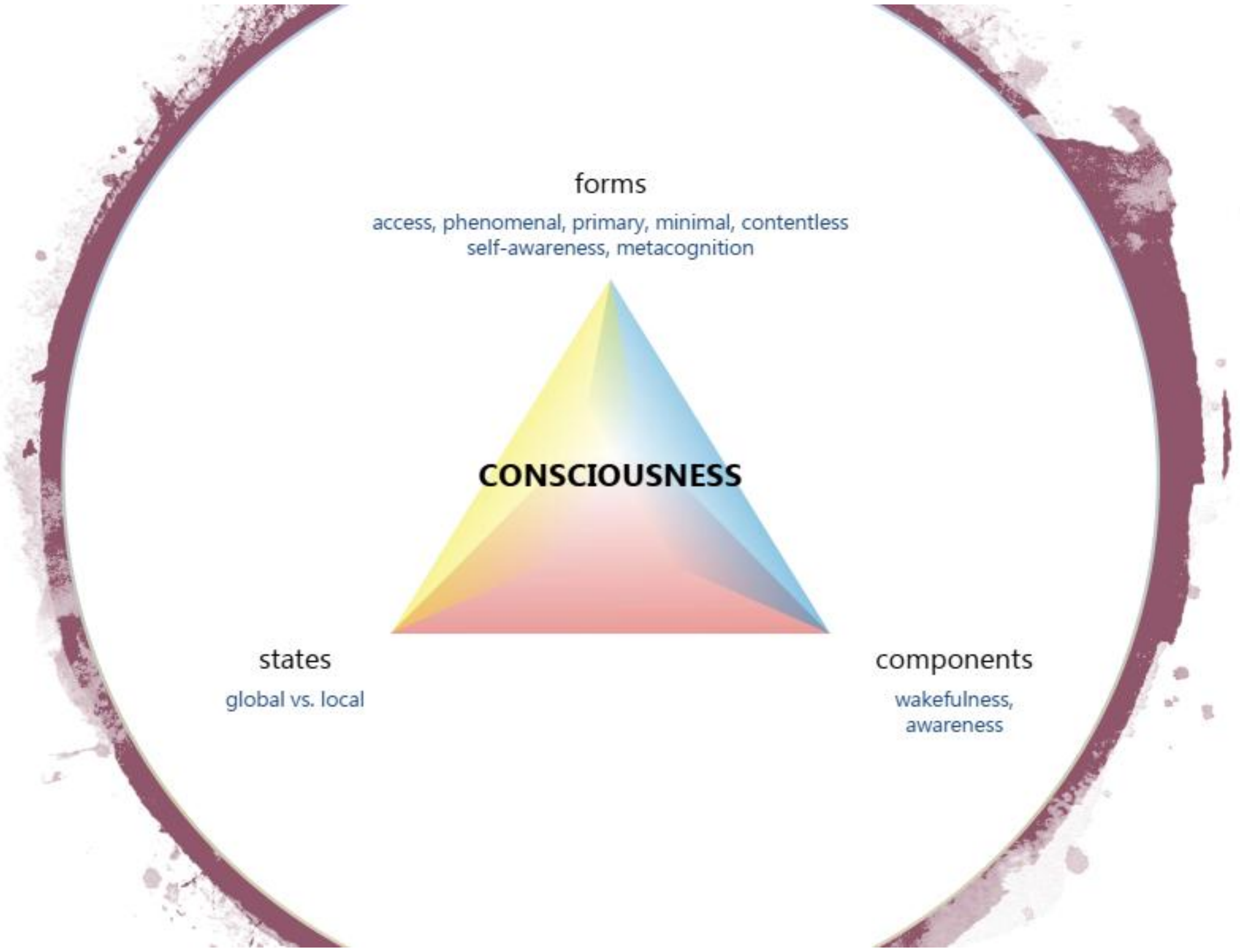

Fig.1. Consciousness as a complex phenomenon has different constituents (states, forms, and components). Each of these constituents has different dimensions.

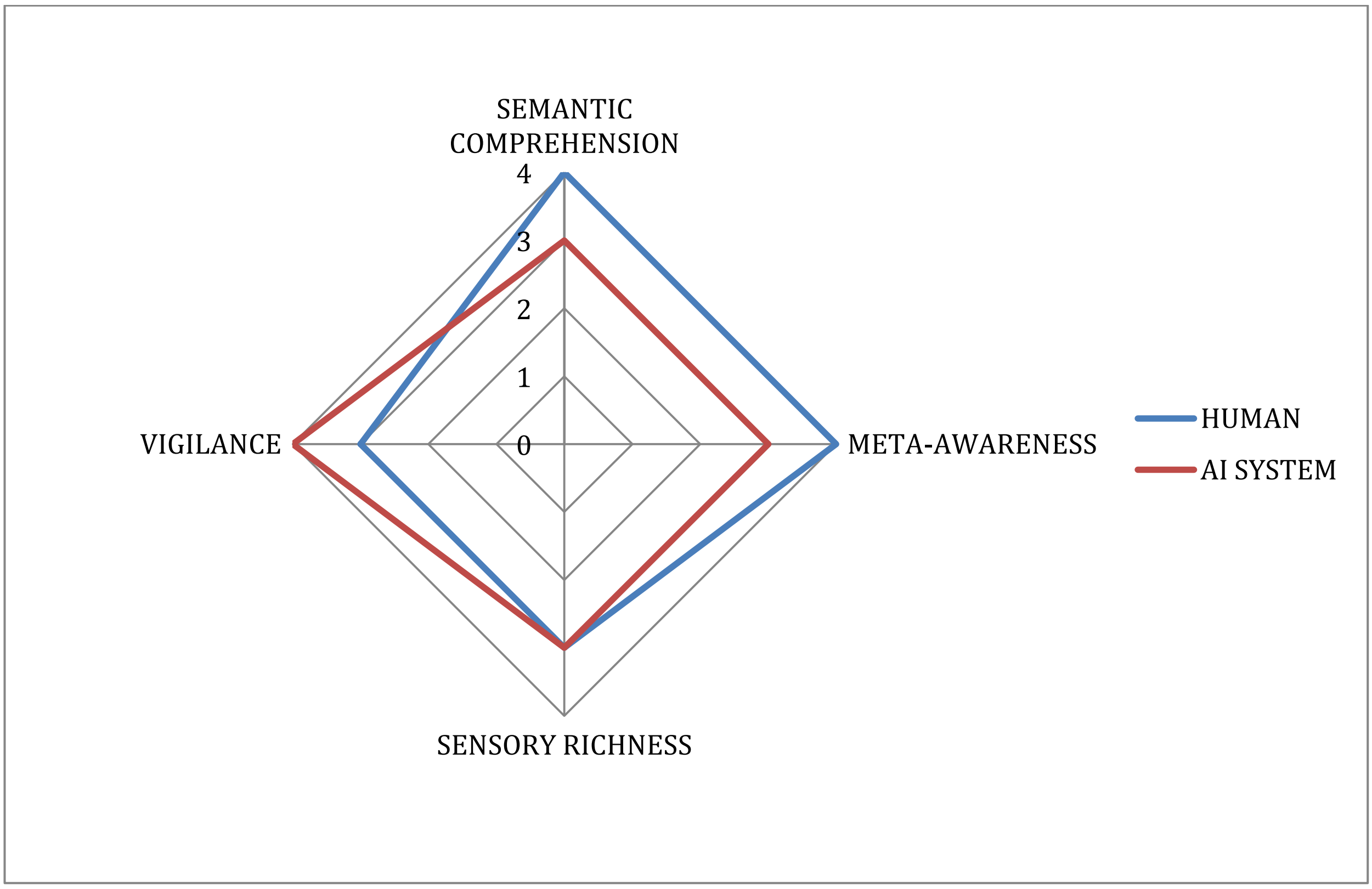

Fig. 2. Illustrative comparison of two hypothetical human and AI system's consciousness profiles. The values assigned are speculative and for the only sake of illustration. The human and AI system's consciousness profiles are represented by the blue and red diamonds respectively. This illustration is based on [86], who apply the same approach to animal consciousness.

## Acknowledgements

This research has received funding from the project Counterfactual Assessment and Valuation for Awareness Architecture—CAVAA (European Commission, EIC 101071178). We thank two anonymous reviewers for their comments which helped us to improve both the readability and the clarity of the text.